\documentclass{article}
\usepackage{ijcai19}

\usepackage{times}
\usepackage{soul}
\usepackage{url}
\usepackage[hidelinks]{hyperref}
\usepackage[utf8]{inputenc}
\usepackage[small]{caption}
\usepackage{graphicx}
\usepackage{subfigure}
\usepackage{amsmath}
\usepackage{booktabs}
\usepackage{color}
\title{Source Dependency-Aware Transformer with Supervised Self-Attention}

\author{
Chengyi Wang$^1$ Shuangzhi Wu$^2$ Shujie Liu$^3$
\affiliations
$^1$Nankai University, Tianjin, China\\
$^2$Harbin Institute of Technology, Harbin, China \\
$^3$Microsoft Research Asia, Beijing, China
\emails
cywang@mail.nankai.edu.cn \{v-shuawu, shujliu\}@microsoft.com}

\begin{document}

\maketitle

\begin{abstract}
Recently, Transformer has achieved the state-of-the-art performance on many machine translation tasks. However, without syntax knowledge explicitly considered in the encoder, incorrect context information that violates the syntax structure may be integrated into source hidden states, leading to erroneous translations. In this paper, we propose a novel method to incorporate source dependencies into the Transformer. Specifically, we adopt the source dependency tree and define two matrices to represent the dependency relations. Based on the matrices, two heads in the multi-head self-attention module are trained in a supervised manner and two extra cross entropy losses are introduced into the training objective function. Under this training objective, the model is trained to learn the source dependency relations directly. Without requiring pre-parsed input during inference, our model can generate better translations with the dependency-aware context information. Experiments on bi-directional Chinese-to-English, English-to-Japanese and English-to-German translation tasks show that our proposed method can significantly improve the Transformer baseline. 
\end{abstract}

\section{Introduction}

The past few years have witnessed the rapid development of neural machine translation (NMT).\cite{DBLP:journals/corr/BahdanauCB14,DBLP:conf/acl/GehringAGD17,DBLP:conf/icml/GehringAGYD17,DBLP:conf/nips/VaswaniSPUJGKP17} 
Particularly, the Transformer has refreshed the state-of-the-art performance on many translation tasks. Different from recurrence and convolution based network structures, the Transformer relies solely on the multi-head self-attention mechanism in which different heads implicitly model the inputs from different aspects \cite{DBLP:conf/nips/VaswaniSPUJGKP17}. 

\begin{figure*}
\centering
\subfigure[translation example]
{
  \label{sentence}
  \includegraphics[height=0.18 \textwidth]{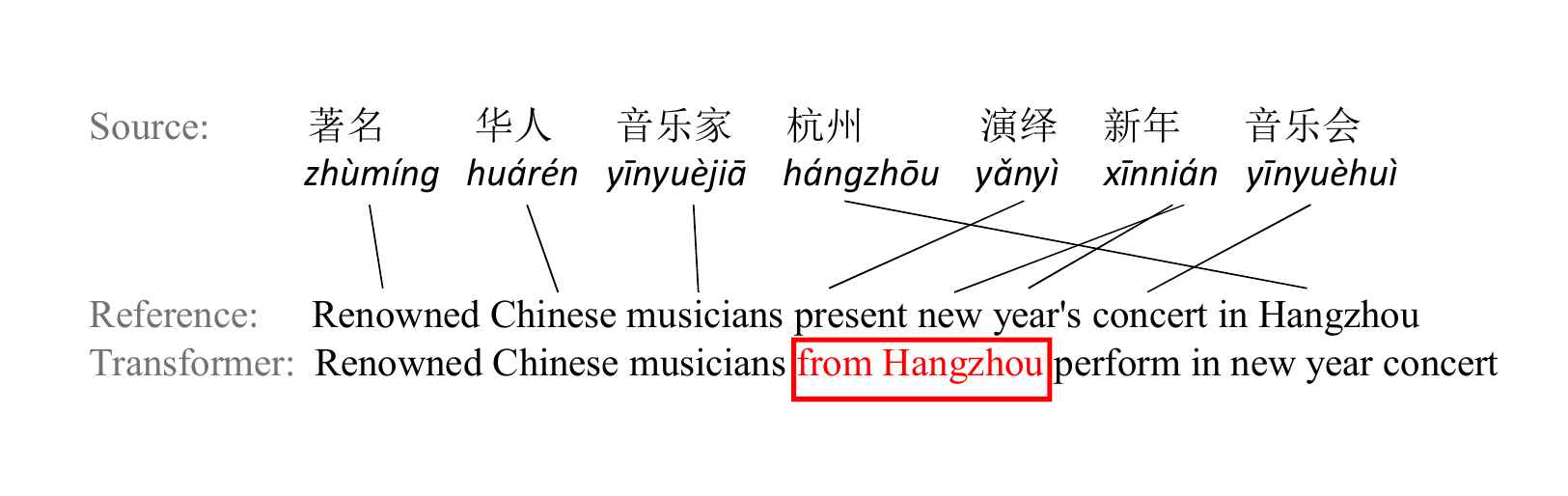}
}
\subfigure[source side dependency tree]
{
  \label{tree}
  \includegraphics[height=0.25\textwidth]{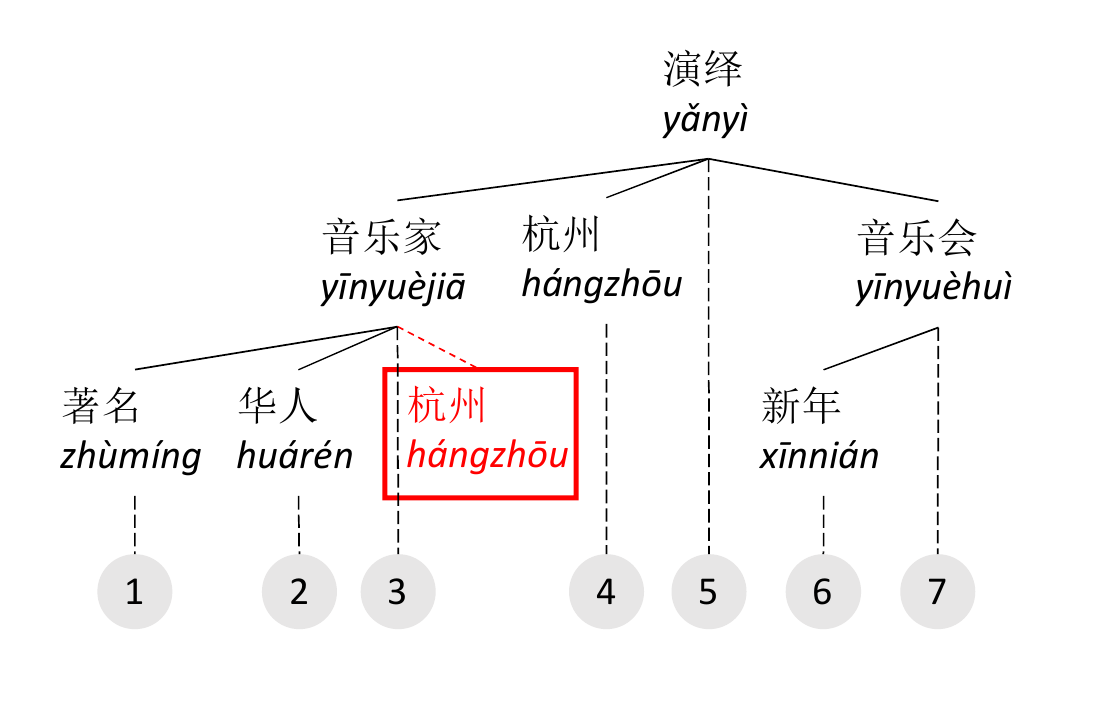}
}
  \caption{(a). A translation example from the Chinese-to-English task. Text highlighted in the rectangle is the incorrect translation part. (b). The dependency tree of the source sentence. The highlighted phrase in the rectangle refers to the Transformer's  misunderstanding of the source sentence.}
\label{example}
\end{figure*}

Although effective, all multi-head self-attentions are trained in an unsupervised manner without any explicit modeling of syntactic knowledge, which leads to incorrect translations that violate the syntactic constraints of source sentences.
Figure \ref{sentence} shows an example from the Chinese-to-English task. Though the translation is well formed and grammatical, its meaning is inconsistent with the source sentence. This error is caused by the misunderstanding of the subtle source syntactic dependency. As shown in Figure \ref{tree}, the word ``\textit{h\'{a}ngzh\={o}u} (Hangzhou)'' is a modifier of ``\textit{y\v{a}ny\`{i}} (present)'' rather than ``\textit{yinyu\`{e}ji\={a}} (musicians)''. Intuitively, such information can be effectively modeled by syntax structure such as dependency trees. Recent advances show that adding source syntactic information to the RNN-based NMT systems can improve translation quality. 
For example, \citeauthor{DBLP:conf/acl/EriguchiHT16} \shortcite{DBLP:conf/acl/EriguchiHT16} and \citeauthor{DBLP:conf/acl/ChenHCC17} \shortcite{DBLP:conf/acl/ChenHCC17} construct a tree-LSTM encoder on top of the standard sequential encoder;
\citeauthor{DBLP:conf/emnlp/BastingsTAMS17} \shortcite{DBLP:conf/emnlp/BastingsTAMS17} introduce an extra graph convolutional network (GCN) to encode dependency trees. 

Though remarkable progress has been achieved, several issues still remain in this research line:

(1) Most existing methods introduce extra modules in addition to the sequential encoder such as the tree-LSTM, GCN or additional RNN encoder, which make the model heavy. 

(2) Existing methods require a stand-alone parser to pre-generate syntactic trees as input during inference, since they are incapable to construct syntactic structures automatically.

(3) Previous methods are designed for RNN-based models and hard to be applied to highly parallelized Transformer. 

To address these issues, we propose a novel framework to enable the Transformer to model source dependencies explicitly by supervising parts of the encoder self-attention heads. 
The self-attention heads are divided into two types: the unsupervised heads and the supervised heads. The unsupervised heads implicitly generalize patterns from raw data as in original Transformer, but the supervised ones learn to align to child and parent dependency words, guided by the parsing tree from a pre-trained dependency parser\footnote{To make the description clear, in the following of this paper, we use ``parent'' to denote the head/parent node of dependency tree and ``head'' to denote the attention head. }.
As the implementation of the supervision, two regularization terms are introduced into the original cross-entropy loss function.
With this method, the dependency information are naturally modeled with several heads without introducing extra modules, and in inference time, the supervised heads can predict the dependency relations without parsed input. 
Experiments conducted on bidirectional Chinese-English, English-to-Japanese and English-to-German show that the proposed method improves translation quality significantly over the strong Transformer baseline and can construct reasonable source dependency structures as well.

Our main contributions are summarized as follows:
\begin{itemize}
\item We propose a novel method to incorporate source dependency knowledge into the Transformer without introducing additional modules. Our approach takes advantage of the numerous attention heads and trains some of them in a supervised manner. 
\item With supervised framework, our method can build reasonable source dependency structures via self-attention heads and extra parsers are not required during inference. Decoding efficiency is therefore guaranteed. 
\item Our proposed method significantly improves the Transformer baseline and outperforms several related methods in four translation tasks.
\end{itemize}

\begin{figure*}
\centering
  \includegraphics[height=0.4\textwidth]{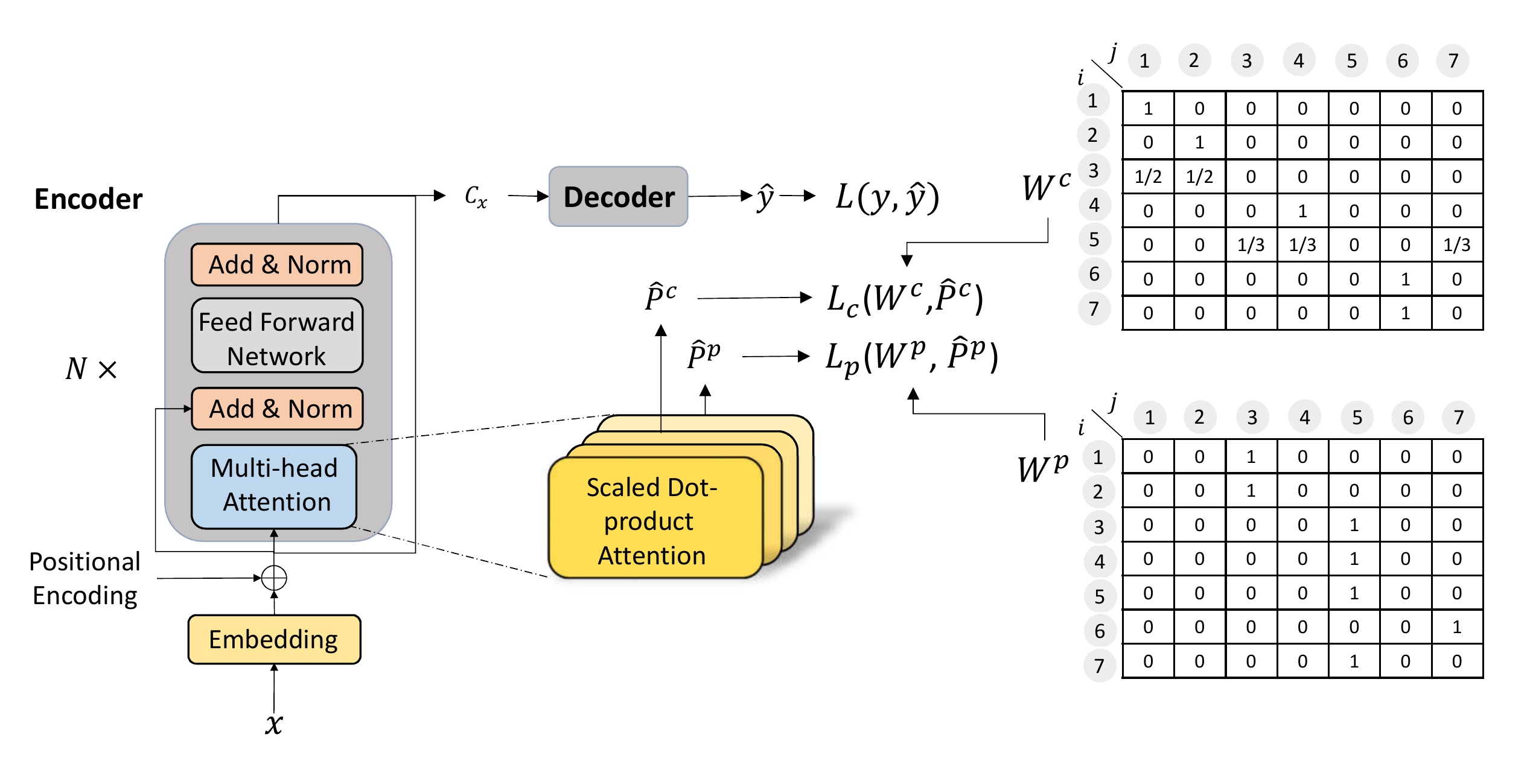}
  \caption{The architecture of our syntax-aware Transformer. $W^c$ and $W^p$ are child attentional adjacency matrix and parent attentional adjacency matrix corresponding to the tree in Figure \ref{tree}, which are used as supervisions of attention score matrices. Each element shows the attention weight upon the $j$-th word based on the $i$-th word.}
  \label{model}
\end{figure*}

\section{Background}
\subsection{Transformer} \label{transformer}
\paragraph{Encoder}Transformer encoder is composed of a stack of $N$ identical layers, where each layer contains a self-attention module and a feed-forward network. 

The self-attention module is designed as a multi-head attention mechanism. The input consists of a query $Q$, a key $K$ and a value $V$. Every elements of $(Q, K, V)$ are modeled as the word and position embedding representation for the first layer and as output of the previous layer for other layers. Multi-head attention linearly projects the $(Q, K, V)$ $h$ times with different learned projections to $d_k$, $d_k$ and $d_v$ dimensions respectively. Then the scaled dot-product attention function performs on them and yields $h$ different $d_v$-dimensional representations. They are then concatenated and projected again to generate the final values:
\begin{small}
\begin{flalign}
\text{MultiHead}(Q,K,V) &= \text{Concat}_i(\text{head}_i)W^O \\
\text{head}_i &= \text{Attention}(QW^Q_i, KW_i^K, VW_i^V) \\
\text{Attention}(Q,K,V) &= \text{softmax}(\frac{QK^T}{\sqrt{d_k}})V 
\label{att_eq}
\end{flalign}
\end{small}where $W_i$ and $W^O$ are learned projection matrices. Note that \(W_i^Q \in R^{d_{model} \times d_k}\), \(W_i^K \in R^{d_{model} \times d_k}\), \(W_i^V \in R^{d_{model} \times d_v}\) and \(W_i^O \in R^{hd_{v} \times d_{model}}\), where $h$ denotes the number of head and $d_k=d_v=d_{model}/h$ in practice. 


The feed-forward network (FFN) is formed as two linear transformations with a ReLU activation in between. 

Layer normalization and residual connection are used after each sub-layer.



\paragraph{Decoder}  Aside from the self-attention and feed-forward module, the decoder inserts an inter multi-head attention sub-layer which performs over the encoder output. Specifically, the output of the self-attention sub-layer is regarded as $Q$ and linearly projected to $d_k$-dimensional queries. The encoder output is regarded as $K$ and $V$, which are projected to $d_k, d_v$ dimensions respectively.

\subsection{Dependency Tree} \label{dependency tree}
A dependency tree directly models syntactic structures of arbitrary distance, where each word has a parent word that it depends on, except for the root word. The verb is taken to be the structural center and all other syntactic units are either directly or indirectly connected to it. 

Figure \ref{tree} shows an example of a dependency tree. 
Without any constituent labels, dependency tree is simple in form but effectively characterizes word relations. Hence, it is usually regarded as a desirable linguistic model of the sentence.

\section{Proposed Method} \label{proposed method}
Figure \ref{model} sketches the overall architecture of our proposed method. We adopt the source dependency tree, and define two attentional adjacency square matrices, $W^c$ and $W^p$, to capture child and parent dependencies. The two matrices are used to supervise two self-attention heads in the Transformer encoder. In this section, we begin by explaining how the source syntax is represented and then give details on how the model is trained based on the learned representations.  

\subsection{Syntax Representation}
Given a source sentence $x = x_1, x_2, ...,x_m$, where $m$ is the sentence length, and its dependency parse tree, we define a child attentional adjacency matrix $W^c \in R^{m \times m}$ and a parent attentional adjacency matrix $W^p \in R^{m \times m}$, representing child and parent dependencies respectively. 

Equation (\ref{Wc}) gives the definition of matrix $W^c$. Assuming that $x_i$ is a possible parent, the element $W^c_{ij}$ is 1 when $x_j$ is a child of $x_i$, otherwise 0. For all leaf nodes in the tree, we let them align to themselves. In cases where a parent node has multiple child nodes, we average the weight among all its child nodes. In this manner, each word is informed with its modifiers directly.

\begin{equation}\small
W^c_{ij} = \left\{
             \begin{array}{ll}
             1/n_i, & x_j\ \mathrm{is\ a\ child\ node} \\
             1, &i=j\ \mathrm{and}\ x_i\ \mathrm{is\ a\ leaf\ node}\\    
             0, &\mathrm{otherwise.}          
             \end{array}
\right.
\label{Wc}
\end{equation}$n_i$ is the number of child nodes of $x_i$. 

Similarly, in $W^p$, each word is encouraged to attend to its parent node directly and the attention score of the parent node is 1. The root node is aligned to itself as shown in Equation (\ref{Wp}).

\begin{equation}\small
W^p_{ij} = \left\{
             \begin{array}{ll}
             1, & x_j\ \mathrm{is\ parent\ node\ or}\ i=j\ \mathrm{and}\ x_i\ \mathrm{is\ root} \\
             0, & \text{otherwise}.          
             \end{array}
\right.
\label{Wp}
\end{equation}

Either of the two matrices is sufficient to reconstruct the dependency tree, and they sketch the tree from different views. Figure \ref{model} gives an example of the attentional adjacency matrices ($W^c$ and $W^p$), which corresponds to the dependency tree in Figure \ref{tree}. In $W^c$, three child nodes of the root word ``\textit{y\v{a}ny\`{i}}'' (the 5th row) receive an equal attention score whereas the others are given no attention. For the parent word ``\textit{y{i}nyu\`{e}hu\`{i}}'' (the 7th row), it only has one child word ``\textit{xinni\'{a}n}'' (the 6th column) and thus, it receives an attention score of 1. As the leaf node, the word ``\textit{xinni\'{a}n}'' (the 6th row) scores itself as  1. In $W^p$, all words put the whole attention score on its parent node except for the root word (the 5th row), which attends to itself. 

\subsection{Syntax-Aware Transformer} \label{syntax-aware transformer}
Inspired by the head selection idea for dependency parsing \cite{zhang2017dependency}, we propose a supervised framework where two self-attention heads in the encoder are supervised by two  attentional adjacency matrices. 
The supervised heads are expected to model the dependency knowledge from different view. 
As shown in Figure \ref{model}, the encoder encodes $x$ and generates a hidden representation $C_x$. The decoder predicts the target sentence $\hat{y}$ based on $C_x$. In the training phase, the objective function is divided into three parts: the standard maximum likelihood of training data and the two regularization terms to encourage the self-attention to learn from adjacency matrices. 

The Transformer encoder contains $N \times h$ self-attention heads for each word in total, where $N$ is the number of layers and $h$ is the number of heads. Though these heads can model the source sentence from different aspects, whether they can learn accurate syntactic knowledge in an unsupervised manner is indecipherable. Thus, we use the attentional adjacency matrices to guide two of the attention heads to explicitly guarantee them to learn syntactic knowledge. 

Specifically, for each source word $x_i$, we expect that the self-attention function can focus more on its child words and parent word. Thus, we select two self-attention heads from the same layer as supervised attention heads and denote them as child supervised attention head (CSH) and parent supervised attention head (PSH). They are trained under the guidance of $W^c$ and $W^p$. The attention score/probability matrices produced by CSH and PSH are denoted as $\hat{P^c}$ and $\hat{P^p}$ respectively. which are computed by softmax functions as in Equation (\ref{att_eq}). Two additional objective terms, namely, $\mathcal{L}_c(W^c, \hat{P}^c)$ and $\mathcal{L}_p(W^p, \hat{P}^p)$ are introduced to minimize the divergence between them and the attentional adjacency matrices $W^c$, $W^p$:

\begin{small}
\begin{equation}
\begin{split}
\mathcal{L}_c(W^c, \hat{P}^c) &= -{W^c}\mathrm{log}\hat{P}^c\\
&= -\sum_{i=1}^{m}\sum_{j=1}^{m}{W^c_{ij}}\mathrm{log}\hat{P}^c_{ij}
\end{split}
\end{equation}
\begin{equation}
\begin{split}
\mathcal{L}_p(W^p, \hat{P}^p) &= -{W^p}\mathrm{log}\hat{P^p}\\
&= -\sum_{i=1}^{m}\sum_{j=1}^{m}{W^p_{ij}}\mathrm{log}\hat{P}^p_{ij}
\end{split}
\end{equation}
\end{small}
The two regularization objectives encourage the CSH and PSH to generate similar intermediate attention weight matrices as attentional adjacency matrices $W^c, W^p$. Other attention heads are still trained in an unsupervised manner as the original Transformer does. 

The original training objective $\mathcal{L}(x,y)$ on source sentence $x$ and target sentence $y$ is defined as:
\begin{equation}
\mathcal{L}(x,y;\theta) =-\mathrm{log}\mathit{P}(y|x;\theta)
\end{equation}

After introducing the regularization terms, the new objective function is formulated as:
\begin{small}
\begin{equation}
\mathcal{J}(\theta) =  \sum_{\langle x,y\rangle \in D}(\mathcal{L}(x,y;\theta) +\alpha \mathcal{L}_c(W^c, \hat{P}^c) +\beta \mathcal{L}_p(W^p, \hat{P}^p))\\
\end{equation}
\end{small}where $\alpha$ and $\beta$ are hyper-parameters for the regularization terms, and $D$ is the training corpus. 

\section{Experiments} \label{experiments}
\begin{table*}[] 
\centering
\begin{tabular}{l|c|c|c|c}\hline
  System             & NIST2005  & NIST2008 & NIST2012 & Average  \\
  \hline
  RNNsearch & 38.41 & 30.01 & 28.48 & 32.30 \\
  Tree2Seq \cite{DBLP:conf/acl/ChenHCC17}  & 39.44 & 31.03 & 29.22 &  33.23\\
  SE-NMT (Wu et al. 2017)     & 40.01  & 31.44 & 29.45 &  33.63\\
  \hline
  Transformer     & 43.89 & 34.83  & 32.59 & 37.10  \\
  $\,\,\,\,\,\,$+CSH     & 44.21 & 36.63  & 33.57 &  38.14\\
  $\,\,\,\,\,\,$+PSH     & 44.24 & 36.17  &  33.86 & 38.09\\
  $\,\,\,\,\,\,$+CSH+PSH    & \bf{44.87} & \bf{36.73} & \bf{34.28} & \bf{38.63}\\\hline

\end{tabular}
\caption{Case-insensitive BLEU scores (\%) for Chinese-to-English translation on NIST datasets. ``+CSH" denotes model only trained under the supervision of child attentional adjacency matrix ($\beta$ =  0). ``+PSH" denotes model only trained under the supervision of parent attentional adjacency matrix ($\alpha$ = 0). ``+CSH+PSH" is trained under the supervision of both. }
  \label{Nist}

\end{table*}
\subsection{Setup} \label{setup}

\paragraph{Dataset:} For NIST OpenMT's Chinese-to-English translation task, we leverage a subset of LDC corpus as bilingual training data, \footnote{LDC2003E14, LDC2005T10, LDC2005E83, LDC2006E26, LDC2003E07, LDC2005T06, LDC2004T07, LDC2004T08, LDC2006E34, LDC2006E85, LDC2006E92, LDC2003E07, LDC2002E18, LDC2005T06} which contains 2.6M sentence pairs. The NIST 2005, 2008, 2012 are used as test sets. All English words are in lowercase. We keep the top 30K most frequent words for both sides, and the rest are replaced with \verb|<unk>| and post-processed following \citeauthor{DBLP:conf/acl/LuongSLVZ15} \shortcite{DBLP:conf/acl/LuongSLVZ15}. 

In the WAT2016 English-to-Japanese translation task, the top 1.5M sentence pairs from the ASPEC corpus \cite{DBLP:conf/lrec/NakazawaYUUSKI16}\footnote{\url{http://orchid.kuee.kyoto-u.ac.jp/ASPEC/}} are used as training data. We follow the official pre-processing steps provided by WAT2016.

For WMT2017's bidirectional Chinese-English translation tasks, we use the CWMT corpus\footnote{\url{http://www.statmt.org/wmt17/translation-task.html}}, which consists of 9M sentence pairs.  The newstest2017 is used as the test set. For pre-processing, we segment Chinese sentences with our in-house tool and segment English sentences with Moses scripts\footnote{\url{https://github.com/moses-smt/mosesdecoder/blob/master/scripts/tokenizer/tokenizer.perl}}. We use 50k subword tokens as vocabulary based on Byte Pair Encoding (BPE) \cite{DBLP:conf/acl/SennrichHB16a} for both sides'.

For English-to-German task, we use the WMT2014 corpus, which contains 4.5M sentence pairs. The newstest 2014 is used as test set. We use vocabularies of 50K sub-word tokens based on BPE for both sides.

Given that no golden annotations of source dependency trees exist in these corpus, we use pseudo parsing results from in-house implemented arc-eager dependency parsers following \citeauthor{DBLP:conf/acl/ZhangN11} \shortcite{DBLP:conf/acl/ZhangN11}. The English parser is trained on the Penn Treebank and the Chinese parser is trained on Chinese Treebank corpus. The unlabeled attachment score (UAS) are 92.3\% and 83.7\% respectively. As for tasks using BPE, we modify the pseudo-golden dependency trees by a rule: all pieces from one word are linked to the first piece.

We compare our proposed method with the Transformer \cite{DBLP:conf/nips/VaswaniSPUJGKP17}. The results are reported with the IBM BLEU-4. The English-to-Japanese task is evaluated following the official procedure with both BLEU and RIBES.

\paragraph{Model and Implementation Details:}
The Transformer baseline and our proposed method follow the base setting of \citeauthor{DBLP:conf/nips/VaswaniSPUJGKP17} \shortcite{DBLP:conf/nips/VaswaniSPUJGKP17}. CHS and PSH are selected from the top layer. Our off-line experiments show that the model exhibits best performance when supervision is conducted on the top layer. {This may be because the syntax information captured by lower layers is weakened as the encoder goes deeper, however, when the supervision is on the top layer, this kind of information is more strong and effective. } The hyper-parameters used in our approach are set as $\alpha = 0.4, \beta = 0.4$, both are selected based on validation set. All experiments are conducted on a single GPU.

\begin{table}[]
\centering
\begin{tabular}{l|l|l} \hline
System            & BLEU  & RIBES \\ \hline
RNNsearch       & 34.83 & 80.92 \\
Eriguchi et al. (2016) & 34.91 & 81.66 \\
\hline
Transformer & 36.24 & 81.90 \\
$\,\,\,\,\,\,$+CSH    & 36.83 & 82.15      \\
$\,\,\,\,\,\,$+PSH    & 36.75 & 82.09      \\
$\,\,\,\,\,\,$+CSH+PSH & \bf{37.22} & \bf{82.37} \\ \hline
\end{tabular}
\caption{Evaluation results on the English-to-Japanese translation task.}
\label{WAT}
\end{table}
\subsection{Evaluation on NIST Chinese-to-English Translation} \label{Cn-En-Exp}
In this experiment, aside from the Transformer, we also compare our model with the RNN-based NMT baseline RNNsearch and several existing syntax-aware NMT methods that use source consistency/dependency trees. These methods are described as follows:
\begin{itemize}
\item RNNSearch: A reimplementation of the conventional RNN-based NMT model \cite{DBLP:journals/corr/BahdanauCB14}.
\item Tree2Seq: \citeauthor{DBLP:conf/acl/ChenHCC17} \shortcite{DBLP:conf/acl/ChenHCC17} propose a tree-to-sequence NMT model by leveraging source constituency trees with tree based coverage.\footnote{\url{https://github.com/howardchenhd/Syntax-awared-NMT}}. 
\item SE-NMT: \citeauthor{DBLP:conf/ijcai/WuZZ17} \shortcite{DBLP:conf/ijcai/WuZZ17} extract extra sequences by traversing and encode them using two RNNs. We re-implement their model named SE-NMT. 
\end{itemize}

Table \ref{Nist} shows the evaluation results for all test sets. We report on case-insensitive BLEU here since English words are lowercased. From the table we can see that syntax-aware RNN models always outperform the RNNsearch baseline. However, the performance of the Transformer is much higher than that of all RNN-based methods.  In Transformer+CSH, we use only the child attentional adjacency matrix to guide the encoder. Being aware of child dependencies, Transformer+CSH gains 1.0 BLEU point improvement over the Transformer baseline on the average. Transformer+PSH, in which only the parent attentional adjacency matrix is used as supervision, also achieves about 1.0 BLEU point improvement. After combination, the new Transformer+CSH+PSH can further improve BLEU by about 0.5 point on the average, which significantly outperforms the baseline and other source syntax-based methods in all test sets. This demonstrates that both child dependencies and parent dependencies benefit the Transformer model and their effects can be accumulated.

\subsection{Evaluation on English-to-Japanese task}
We conduct experiments on the WAT2016 English-to-Japanese translation task in this section. Our baseline systems include RNNsearch, a tree2seq attentional NMT model using tree-LSTM proposed by Eriguchi et al. (2016) and Transformer. 
Table \ref{WAT} shows the results. According to the table, our Transformer+CSH and Transformer+PSH outperform Transformer and the other existing NMT models in terms of both BLEU and RIBES. Similar as we get in Section \ref{Cn-En-Exp}, the Transformer+CSH+PSH gets the highest performance.

\begin{table} 
\centering
\begin{tabular}{l|c|c|c|c}
\hline
           			& {Zh-En} & \multicolumn{2}{c|}{En-Zh} & {En-De} \\ \cline{3-4}
  System   			&   &   CBLEU  & WBLEU & \\ \hline
 Transformer  & 21.29  & 32.12 & 19.14 & 25.71\\ 
 $\,\,\,\,$+CSH  & 21.60 & 32.46  & 19.54 & 26.01 \\
 $\,\,\,\,$+PSH  & 21.67 & 32.37  & 19.53 & 25.87 \\
 $\,\,\,\,$+CSH+PSH & \bf{22.15} & \bf{33.03}  & \bf{20.19} & \bf{26.31} \\ \hline
\end{tabular}
\caption{BLEU scores (\%) for Chinese-to-English (Zh-En), English-to-Chinese (En-Zh) translation on WMT2017 datasets and English-to-German (En-De) task. Both char-level BLEU (CBLEU) and word-level BLEU (WBLEU) are used as metrics for the En-Zh task.}
\label{WMT}
\end{table}

\subsection{Evaluation on the WMT Tasks}
To verify the effect of syntax knowledge on large-scale translation tasks, we further conduct three experiments on the WMT2017 bidirectional English-Chinese tasks and WMT2014 English-to-German. The results are listed in Table \ref{WMT}. For the Chinese-to-English, our proposed method outperforms baseline by 0.86 BLEU score. For the English-to-Chinese task, the Transformer+CSH+PSH gains 0.91 and 1.05 improvements on char-level BLEU and word-level BLEU respectively. For En-De task, the improvement is 0.6 which is not as much as the other two. We speculate that as the grammars of English and German are very similar, the original model can capture the syntactic knowledge well. Even though, the improvement still illustrates the effectiveness of our method.


\begin{figure*}
\centering
\subfigure[]
{
  \label{translation case}
  \includegraphics[height=0.19 \textwidth]{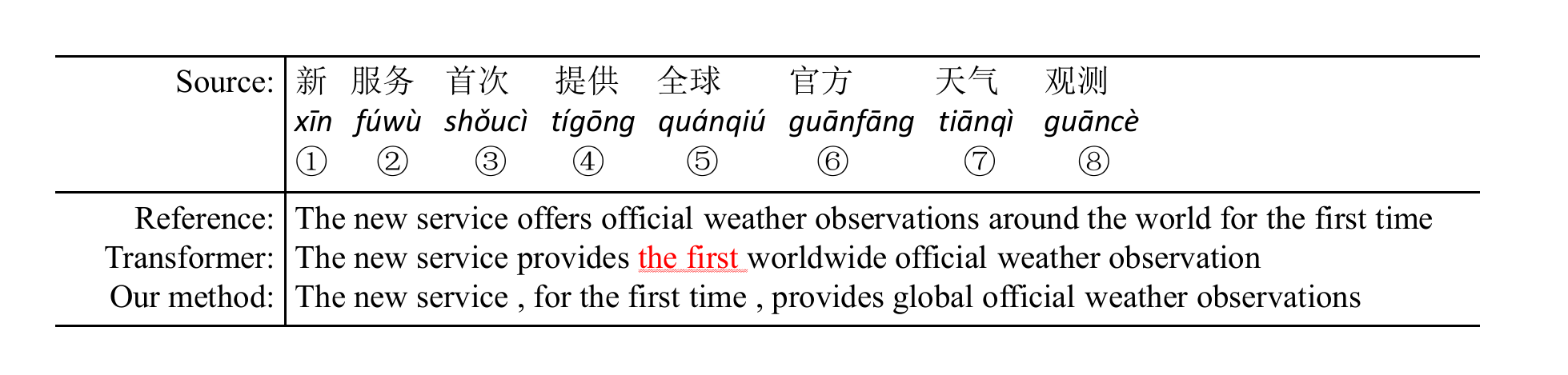}
}
\subfigure[]
{
  \label{case attention}
  \includegraphics[height=0.24 \textwidth]{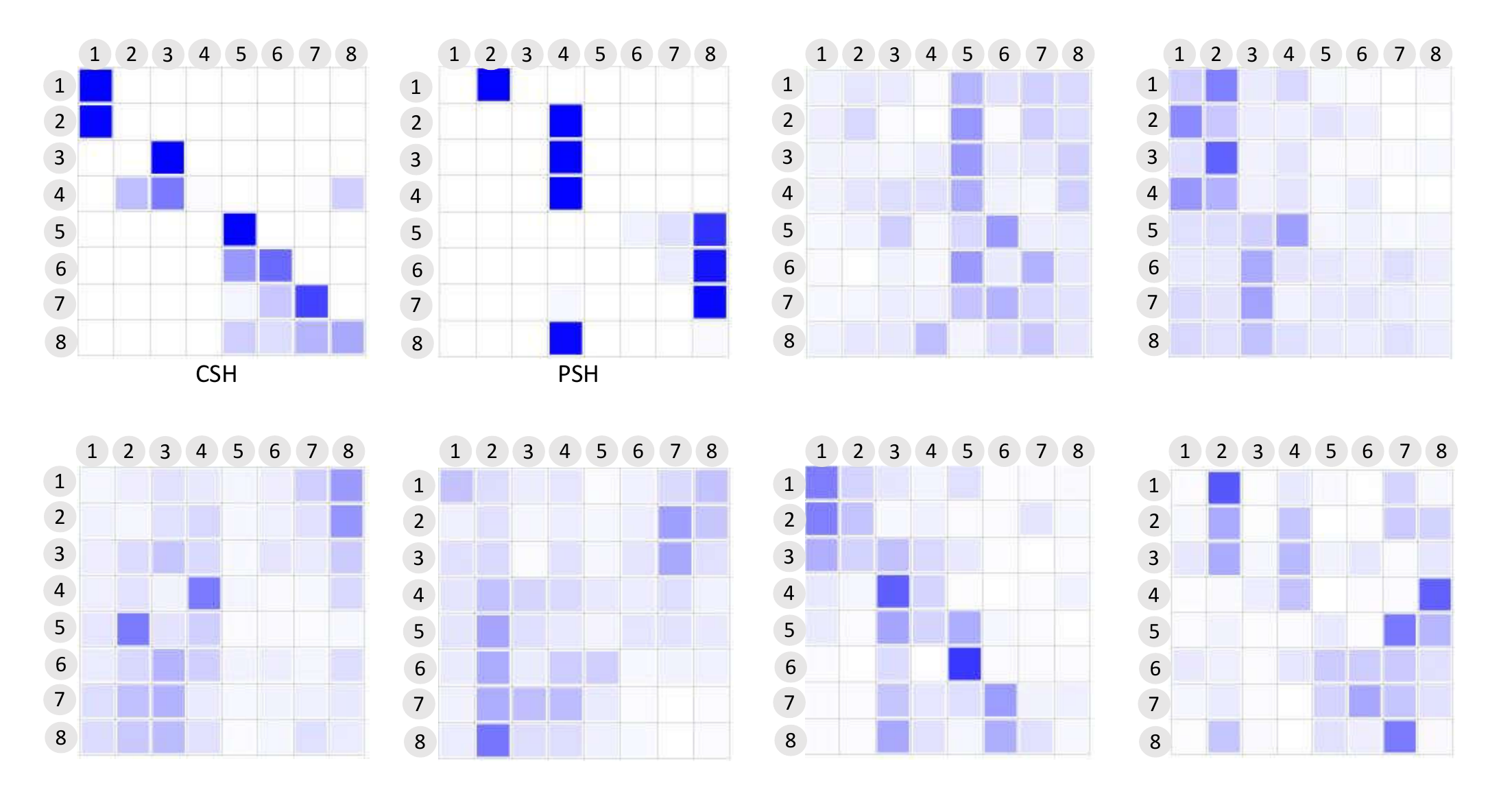}
}  
\subfigure[]
{
  \label{case tree}
  \includegraphics[height=0.24\textwidth]{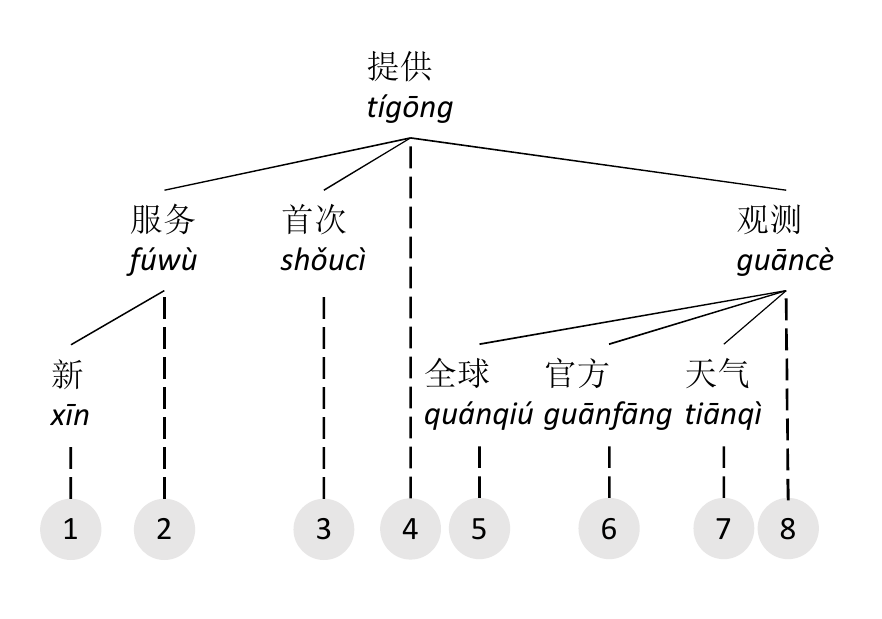}
}
\caption{(a) Translation example from NIST Chinese-to-English task, in which incorrectly translated part is highlighted in wavy line.  (b). Alignments of attention heads in the top layer. The first two are extracted from CSH and PSH respectively. Others are from unsupervised attention heads. Each pixel shows the attention weight. (0: white, 1: blue). (c) The constructed dependency tree based on alignment result of PSH in (b).}
  \label{case study}
\end{figure*}
\subsection{Quality Estimation of Source Dependency Tree Construction}
As CSH and PSH learn distributions over possible parent and child nodes , we can use them to construct source-side dependency trees. In this section, we estimate tree qualities. We only take advantage of the alignment result of PSH since the number of child nodes is uncertain for each word. Specifically, for each word, we denote the node with highest attention weight as the prediction of its parent. The non-tree outputs are adjusted with a maximum spanning tree algorithm 
\cite{chu1965shortest}. We estimate the consistency between the predicted trees and the parsing results of our stand-alone dependency parser due to the unavailable golden references. The higher the consistency is, the closer the performances are. This experiments are preformed on NIST Chinese-to-English task because it has  sufficient test sets. To reduce the influence of ill-formed data as much as possible, we build the evaluation dataset by heuristically selecting 1000 source sentences from all NIST Chinese-to-English testsets that do not contain \verb|<unk>| and have a length of 20-30. Then the parsing results from our stand-alone Chinese parser are used as references. We obtain a UAS of 83.25\%, which demonstrates that the predicted dependency trees are highly similar to the parsing results from the stand-alone parser (the UAS of our stand-alone Chinese parser is 83.7\%). 
\subsection{Case Study}
In this section, we give a case study to explain how our method works. Figure 4(a) provides a translation example from NIST Chinese-to-English test set.   In this example, the Chinese word ``\textit{sh\v{o}uc\`{i}} (the first time)'' (the 3rd word) should be modifier of ``\textit{t\'{i}g\={o}ng} (provides)'' (the 4th word) . However, the Transformer misunderstands source syntax structure and thus generates an incorrect translation. While modeling the source syntax, our proposed model produces a high-quality translation. To further investigate the translation behavior, we visualize the attention weights of different attention heads and show them in Figure 4(b). The first two alignments are extracted from CSH and PSH while the others are from unsupervised ones in the same layer. Different from the general attention heads, CSH and PSH generate more interpretable alignments, based on which we construct the source dependency tree shown in Figure 4(c). From the tree we can see that the dependency of the word ``\textit{sh\v{o}uc\`{i}} (the first time)'' is correctly modeled.

\section{Related Work}

A large body of work has dedicated to incorporating source syntactic knowledge into the RNN-based NMT model. \citeauthor{DBLP:conf/wmt/SennrichH16} \shortcite{DBLP:conf/wmt/SennrichH16} generalize the embedding layer to incorporate morphological features, part-of-speech tags and syntactic dependency labels, leading to improvements on several laguage pairs. \citeauthor{DBLP:conf/acl/EriguchiHT16} \shortcite{DBLP:conf/acl/EriguchiHT16} propose a tree-to-sequence attentional NMT model in which a tree-LSTM is used to encode source-side parse tree.  \citeauthor{DBLP:conf/emnlp/BastingsTAMS17} \shortcite{DBLP:conf/emnlp/BastingsTAMS17} rely on the graph convolutional network (GCN) to incorporate source syntactic structure.  However, they all specify extra features or introduce extra complicated modules in addition to the original sequential encoder. Instead, we focus on the dependency structure and let the model learn from the tree automatically. Other works use linearized representation of parses. For example, \citeauthor{DBLP:conf/acl/LiXTZZZ17} \shortcite{DBLP:conf/acl/LiXTZZZ17} linearize the parse tree of source sentence and use three encoders to incorporate source syntax. \citeauthor{DBLP:conf/ijcai/WuZZ17} \shortcite{DBLP:conf/ijcai/WuZZ17} propose a syntax-aware encoder to enrich each source state with global dependency structure. Though the linearized parses can inject syntactic information into the model without significant changes to the architecture, they are usually lead to much longer input and require additional encoders. 

All these methods are designed for the RNN models and applying them to the highly parallelized Transformer is difficult. Besides, the extra modules to model the source syntax are always heavy and will fail when the input is not parsed. 

\section{Conclusion and Future Work}
In this paper, we propose a novel supervised approach to leverage source dependency tree explicitly into Transformer. Our method is simple and efficient because no extra module is needed and no parser is required during inference. 
Experiments on several translation tasks show that our method yields improvements over the state-of-the-art Transformer model and outperforms other syntax-aware models.


In future work, we expect to achieve developments that will shed more light on utilizing source linguistic features, e.g., dependency labels or part-of-speech tags. Besides, we would like to explore whether incorporating target syntax to Transformer has the potential to improve translation quality.

\bibliographystyle{named}
\bibliography{ijcai19}
\end{document}